\documentclass{article}

\PassOptionsToPackage{numbers}{natbib}  
\usepackage[preprint]{arxiv}
\usepackage[utf8]{inputenc} 
\usepackage[T1]{fontenc}    
\usepackage{hyperref}       
\usepackage{url}            
\usepackage{booktabs}       
\usepackage{amsfonts}       
\usepackage{nicefrac}       
\usepackage{microtype}      
\usepackage{xcolor}         
\usepackage[pdftex]{graphicx}
\usepackage{array} 
\usepackage{float}
\usepackage{natbib}

\title{LA-RCS: LLM-Agent Based Robot Control System}

\author{
  Taek-Hyun Park\\
  Pusan National\\ \ University\\
  \texttt{pthpark1@pusan.ac.kr} \\
  \and
  Young-Jun Choi \\
  National Korea Maritime\\ \& Ocean University \\
  \texttt{akskrunosk0819@gmail.com} \\
  \and
  Seung-Hoon Shin \\
  Republic of Korea Navy \\
  \texttt{tlstndgbs1302@gmail.com} \\
  \and
  Kwangil Lee\thanks{Corresponding author} \\
  National Korea Maritime\\ \& Ocean University\\
  \texttt{leeki@kmou.ac.kr} \\
}

\begin{document}

\maketitle

\begin{abstract}
  \textbf{LA-RCS}(LLM-agent-based robot control system) is a sophisticated robot control system designed to autonomously plan, work, and analyze the external environment based on user requirements by utilizing LLM-Agent. Utilizing a dual-agent framework, LA-RCS generates plans based on user requests, observes the external environment, executes the plans, and modifies the plans as needed to adapt to changes in the external conditions. Additionally, LA-RCS interprets natural language commands by the user and converts them into commands compatible with the robot interface so that the robot can execute tasks and meet user requests properly. During his process, the system autonomously evaluates observation results, provides feedback on the tasks, and executes commands based on real-time environmental monitoring, significantly reducing the need for user intervention in fulfilling requests. We categorized the scenarios that LA-RCS needs to perform into four distinct types and conducted a quantitative assessment of its performance in each scenario. The results showed an average success rate of 90\%, demonstrating the system’s capability to fulfill user requests satisfactorily. For more extensive results, readers can visit our project page: \textcolor{red}{\url{https://la-rcs.github.io/}}

\end{abstract}

\section{Introduction}
The emergence of Large Lagauge Models (LLMs) has shown potential in the fields of complex problems solving, multi-stage inference, and natural language understanding\cite{openai2023gpt4}, especially in generalization performance in various tasks \cite{wei2022emergent} \cite{naveed2023comprehensive} and excellent performance in natural language processing tasks such as understanding, generating, and translating text \cite{brown2020language} \cite{ouyang2022training} \cite{kojima2022large}. Furthermore, as the size of the model increases, LLM can also exhibit novel emergent behaviors, making them useful for solving complex problems or tasks requiring creativity. \cite{zheng2024harnessing} \cite{majumder2024exploring} These large language models evolve to become closer to human thinking, they are able to perform instructions more accurately and consistently in response to user requests \cite{ouyang2022training} \cite{sun2023adaplanner}, and LLM are demonstrating outstanding abilities in everyday life, such as helping humans with various tasks such as translation, summarization, question answering, and creation. \cite{brown2020language} 

Along with the expansion of LLM, there is a research endeavor of the Visual Large Language Models(VLM) \cite{chen2019uniter} \cite{yang2023setofmark} \cite{alayrac2022flamingo}. VLM simultaneously learns the correlation between image and text data through a combination of vision models and language models, and provides enhanced multimodality in response to user requests through its proficient processing capabilities for text and image data.\cite{chen2022multimodal} \cite{yang2023dawn} \cite{zhang2023gpt4v} Multimodal Visual LLM observes user interface (UI) or graphical user interface (GUI) based on image interpretation capabilities, and efficiently performs tasks related to complex user requests and repetitive tasks expressed in natural language using visual information through interaction with user.\cite{zhang2024ufo} \cite{zheng2024gpt4v}

The advancement of Large Language Models (LLM) and related technologies has led to the emergence of LLM-based agents capable of providing problem-solving capabilities across a diverse range of environments\cite{guo2024large} \cite{schick2023toolformer} LLM agents possess the ability to make intelligent decisions, enabling them to plan and reason effectively in response to natural language-based requests\cite{mei2024aios} \cite{shinn2023reflexion} \cite{deng2023mind2web} These technological advances facilitate and provide an important foundation for the evolution of large language models(LLM) into large action models(LAM). LAM control other systems through communication with user and affect the real world through the physical behavior of systems to satisfy user requests.

Previous research in the field of robot control required a large amount of learning data to operate the robot system according to user requests. \cite{zhang2024navid} In addition, there are limitations(complexity, human intervention error ets)of robot system in making arbitrarily automated judgments to fulfill user requests with versatile environments \cite{patil2023Advances} \cite{wake2023gpt4v} and there are endeavors to resolve these limitations\cite{wang2024largelanguage} \cite{liu2024enhancing} \cite{liu2023llmbased}.
In order to fulfill user requests, it is a difficult task to establish and operate a more appropriate plan similar to humans in a situation where various external environments and control objects are faced, rather than in one limited, specific situation. Many steps are required in the decision-making process, such as reasoning and planning, until the action is executed, and it is important to resolve this complexity.

Based on the above challenges, we propose a LLM-Agent-based robot control system (LA-RCS) that minimizes human intervention and complexly considers user requests including the dynamic object(CAROBO)’s surrounding environment. In order to minimize human intervention and take appropriate actions to satisfy user requests, our designed robot control system LA-RCS performs several repetitive functional procedures such as monitoring, observation, planning, and feedback processes based on dual agents (host agent and app agent) to observe the external environment and feedback own action until the user request is completely fulfilled, and constructs control commands appropriate to the user request. In addition, to execute commands more appropriately according to the CAROBO’s external environment from the user request, we used the LLM-Agent with several shot prompting methods. Through the system configuration, finally LA-RCS has strong performance, enhanced autonomy, continuous perform and appropriate command inference process for a given intricated user request based on the external environment of CAROBO.

To evaluate the LA-RCS’s performance rate based on robot(CAROBO)’s external information for user requests and efficiency of it’s autonomous ability in condition of using two different LLM-Agent(GPT-4-Turbo, GPT-4), we considered various situation cases of user requests in testing. The various situations are constructed from four cases and evaluation are conducted on five queries per each cases with different LLM-Agent. The result of evaluation about various testing situations involved quantitative figures, emphasizing the versatility, autonomy and expandability of robot control system in terms of  performing  user intricate requests successfully. 

Our LA-RCS is an sofisticated automation system producing optimized result and considering the robot’s external circumstance from user’s intricate requests and considering the robot’s circumstance with minimizing human intervention in control decision process. Also our designed system is appropriate for versatile situation that require automation and will be applied to many applications in other research. We anticipate that this study will act as a valuable resource for the robotics research community and stimulate further advancements in this field.

\section{Related Work}

\subsection{LLM Agents}
LLM-based autonomous agents have demonstrated remarkable capabilities in reasoning, utilizing tools, and adapting to new observations across various real-world tasks.\cite{yao2022react}\cite{xi2023rise}\cite{wang2023survey} These agents enable LLMs to perform more complex operations compared to traditional LLMs by leveraging human-like decision-making systems.

AutoGPT\cite{autogpt} , a pioneering agent, processes user requests by decomposing the actions of LLMs into thoughts, reasoning, and critiques. TaskWeaver\cite{qiao2023taskweaver} is a code-centric agent framework that converts user requests into executable sub-tasks through Python code. LangChain\cite{chase2022langchain} , another LLM Agent Framework, assists LLMs in utilizing a variety of custom tools, such as Retrieval-Augmented Generation and Chain-of-Thought reasoning.

Utilizing these tools, agents can solve a wide range of problems, including manipulating objects\cite{fan2022minedojo} , controlling GUI \cite{wang2024mobileagent}\cite{yang2023appagent} , and calling API\cite{ge2023openagi}\cite{tang2023toolalpaca}. Multi-agent LLM, which facilitate multi-agent conversations, have emerged as frameworks capable of solving more powerful and complex problems. This architecture effectively assigns tasks to individual agents with specific strengths and enables collaboration or competition among agents to handle complex tasks efficiently.

AutoGen\cite{wu2023autogen} proposes a framework that customizes each agent and utilizes conversations between agents to leverage their specific strengths, thereby enabling a multi-agent system to address user requests effectively. Similarly, frameworks like CrewAI\cite{crewAI}, MetaGPT\cite{hong2023metagpt}, LangGraph\cite{langgraph} , and AutoAgent\cite{chen2023autoagents}  build systems where multiple agents, each with unique roles, can seamlessly perform complex tasks.  

UFO employs a dual-agent framework to meticulously observe and analyze the graphical user interface (GUI) and control information of Windows applications, further highlighting the diverse applications of multi-agent LLM frameworks in real-world scenarios.

\subsection{LLM Agents for Robot Task Planning}
The application of multimodal Large Language Models for robot task planning has been gaining significant traction recently.\cite{mower2023rosllm}\cite{kwon2023trajectory}\cite{ahn2022grounding}\cite{brohan2023rt2}  Previous studies have often utilized few-shot or zero-shot methods for establishing task plans for robots\cite{brown2020fewshot}. Various techniques have emerged for employing LLMs in robot task planning.

For instance, Text2Motion\cite{lin2023text2motion} and Do As I Can, Not As I Say \cite{ahn2022grounding} have demonstrated the use of value functions to enable LLMs to perform task planning for robots. Approaches like RT-2, ChatGPT for Robotics\cite{vemprala2023chatgpt} , and ProgPrompt\cite{singh2022progprompt}  have shown that robot task planning can be effectively achieved through the use of few-shot prompting, chain-of-thought reasoning\cite{wei2022chainofthought} , and the ReAct framework .

Moreover, multi-agent LLM systems have also been applied to robot task planning, showing substantial improvements in performance. SMART-LLM , for example, converts high-level task instructions into multi-robot task plans, allowing multiple robots to collaboratively execute complex tasks. Each step in SMART-LLM\cite{kannan2023smartllm} utilizes LLM prompts to efficiently decompose tasks, form teams, and allocate tasks for execution.

However, existing approaches have primarily focused on task decomposition and teaming within the task planning domain, without addressing the interpretation of high-level commands, situational analysis, and controlling robots through feedback mechanisms.

We focus on using the dual-agent framework proposed in UFO to decompose high-level task commands for robot interfaces. This framework emphasizes the hierarchy of commands performed by each agent, enhancing their ability to interpret and execute commands through communication.

This method enables task planning to adapt in real-time to changing environments, allowing robots to analyze the environment while remaining focused on user commands.

\section{The Design of LA-RCS}

\begin{figure}[htbp]
    \centering
    \includegraphics[width=\textwidth]{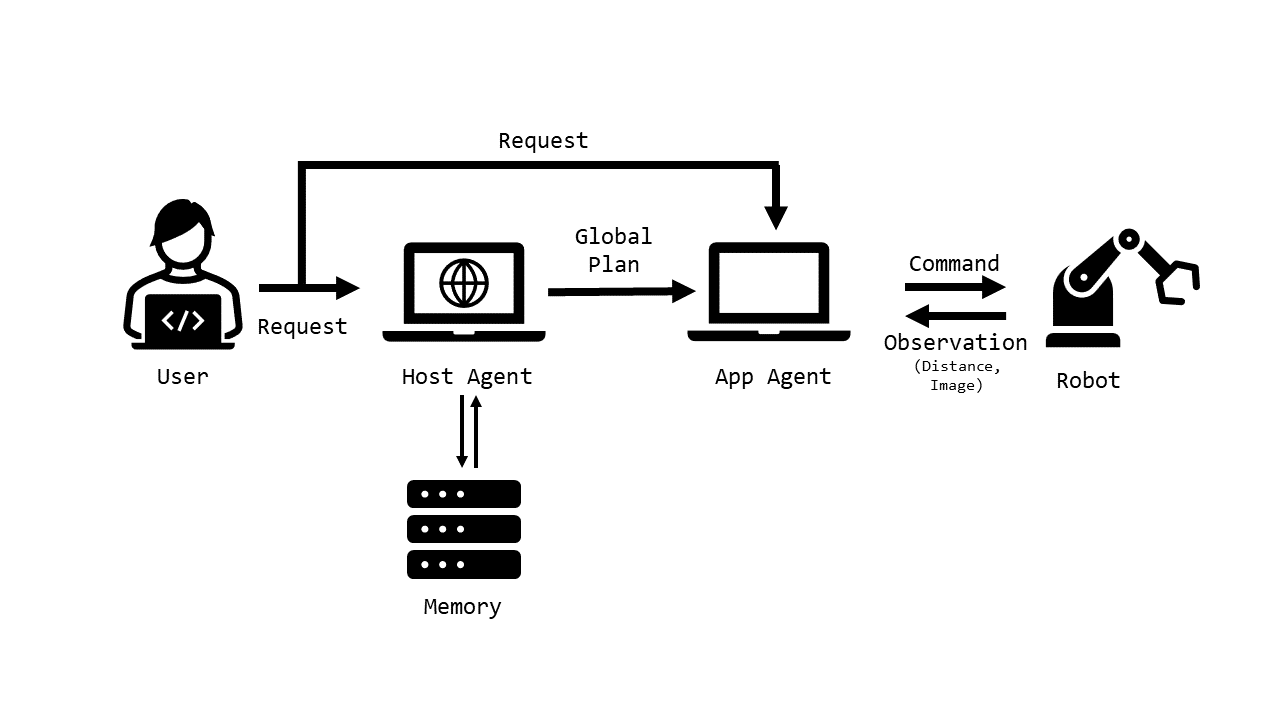}
    \caption{The overall architecture of the LA-RCS.}
    \label{Overall}
\end{figure}

The Figure\ref{Overall} above illustrates the overall structure of LA-RCS. LA-RCS operates on a Dual-Agent framework, comprising two key components:
\begin{itemize}
\item\textbf{Host Agent}: This agent is responsible for constructing a Global Plan to execute the User Request.
\item\textbf{App Agent}: This agent receives the Global Plan from the Host Agent and iteratively performs tasks to fulfill the User Request based on this plan.
\end{itemize}

Both agents utilize GPT-4-Turbo or GPT-4 models to comprehend Observations and execute User Requests. They control CAROBO through self-determined actions using Control Functions (commands).

Upon receiving a User Request, the Host Agent first analyzes the request. LA-RCS provides the Host Agent with Observations and Sensor Data to facilitate the robot's understanding of its current state. Based on this information, the Host Agent constructs a Global Plan, which is then transmitted to the App Agent.

Once the Global Plan is established, the App Agent begins to execute the User Request. The App Agent's decision-making process is informed not only by the received Global Plan but also by Observations, Sensor Data, and Memory. The App Agent selects appropriate Control Functions (Commands) to manipulate the robot and transmits these to the system. This process, performed by the App Agent, is repeated until it determines that the User Request has been successfully completed.

The following sections will delve into more detailed aspects of each component within the LA-RCS framework.

\subsection{Host Agent}

\begin{figure}[!h]
    \centering
    \includegraphics[width=\textwidth]{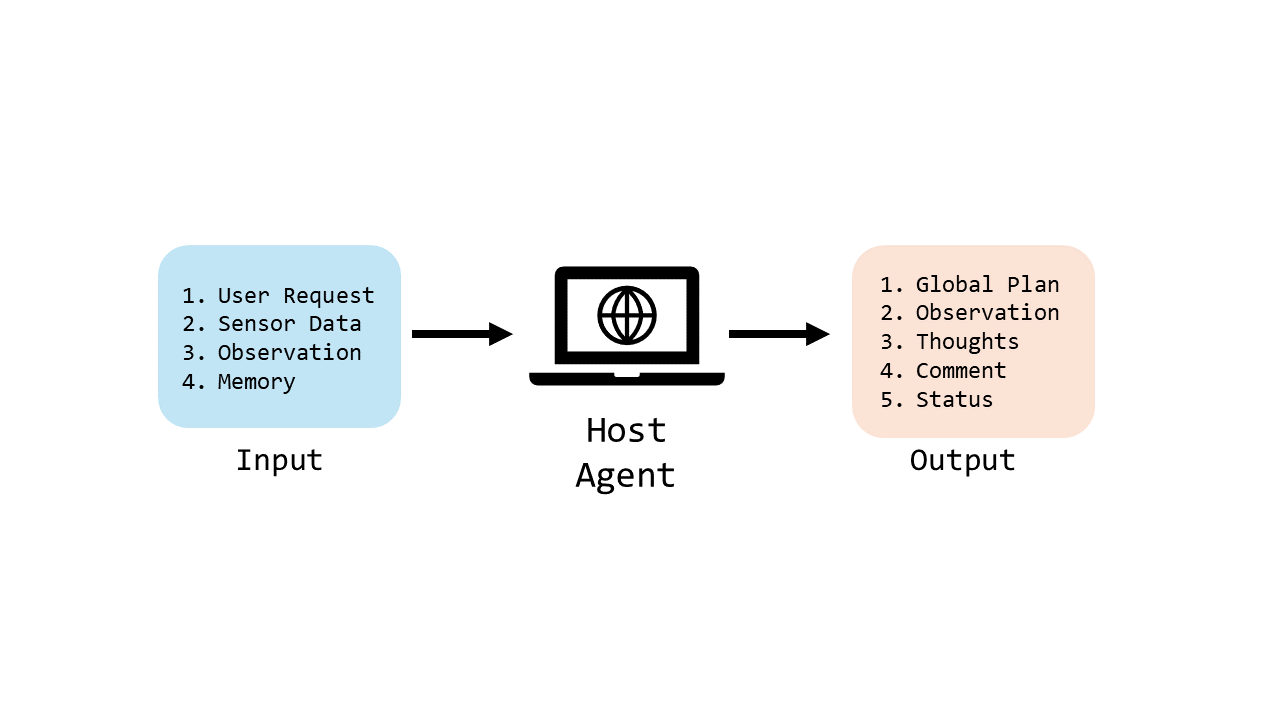}
    \caption{An illustration of the HostAgent.}
    \label{HostAgent}
\end{figure}

\subsubsection{Host Agent Input}
\begin{itemize}
    \item \textbf{User Request}: The robotic action requested by the user.
    \item \textbf{Sensor Data}: Data from the robot's sensors.
    \item \textbf{Observation}: Visual data from the robot's camera.
    \item \textbf{Memory}: Previous thoughts, comments, actions, and execution results.
\end{itemize}

Through Vision Data and Sensor Data, the Host Agent comprehends the current state and can constrain the selection of Control Functions for robot movement. Additionally, Memory provides past experiences such as previous comments and actions, enabling decision-making based on this information. This diverse input framework enhances the Host Agent's capability to fulfill User Requests.

\subsubsection{Host Agent Output}
Utilizing all input information, the Host Agent employs GPT-4o to generate the following outputs:
\begin{itemize}
    \item \textbf{Global Plan}: Action plan to fulfill the user's request.
    \item \textbf{Observation}: Detailed description of Vision Data.
    \item \textbf{Thoughts}: Logical next steps required to meet the given task.
    \item \textbf{Comment}: Progress status and information to be provided.
\end{itemize}

The Host Agent generates various outputs for the following reasons:
\begin{enumerate}
    \item To ensure clear analysis of the current situation in constructing the Global Plan and to provide the process and logic behind decisions.
    \item To explain progress to the User or answer User inquiries.
\end{enumerate}

\subsection{App Agent}

\begin{figure}[htbp]
    \centering
    \includegraphics[width=\textwidth]{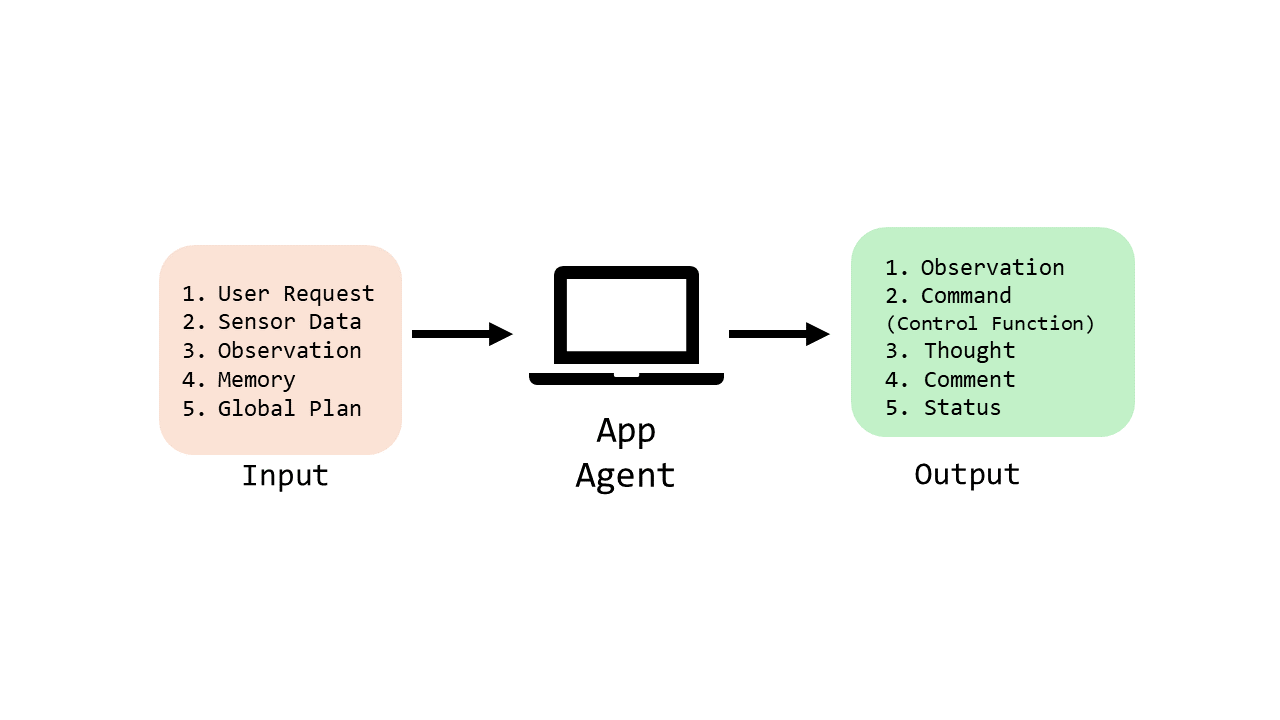}
    \caption{ An illustration of the AppAgent.}
    \label{AppAgent}
\end{figure}

\subsubsection{App Agent Input}
\begin{itemize}
    \item \textbf{User Request}: The robotic action requested by the user.
    \item \textbf{Global Plan}: Action plan to fulfill the user's request.
    \item \textbf{Observation}: Visual data from the robot's camera.
    \item \textbf{Memory}: Previous thoughts, comments, actions, and execution results.
    \item \textbf{Sensor Data}: Data for determining action initiation or termination based on the distance between the robot and objects.
\end{itemize}

As previously explained, the App Agent determines the robot's actions based on the Global Plan to fulfill the User Request. Memory provides the robot's past actions, allowing the App Agent to analyze and reduce the probability of repeating inefficient or meaningless actions.

\subsubsection{App Agent Output}
Based on these inputs, the App Agent analyzes the information and produces the following outputs:
\begin{itemize}
    \item \textbf{Comment}: Progress status and information to be provided.
    \item \textbf{Control Function (Command)}: Action to be performed by the robot.
    \item \textbf{Observation}: Analyzed Vision Data.
    \item \textbf{Status}: Task status, \texttt{"CONTINUE"} if additional action is needed, \texttt{"FINISH"} if the action is completed.
\end{itemize}

The App Agent determines the next step based on these output states. It repeatedly performs observation through Vision Data and robot control until the action is completed, i.e., until the Status becomes \texttt{FINISH}. These outputs are continuously stored in Memory, contributing to the App Agent's correct decision-making in fulfilling User Requests.

\subsection{CAROBO}
We constructed CAROBO as a robot platform to implement LA-RCS. CAROBO is a car-type robot built using the Raspberry Pi 4B model. To provide Vision Data to LA-RCS, CAROBO incorporates a camera. For Sensor Data, it includes ultrasonic sensors and infrared obstacle avoidance sensors. Additionally, we integrated motors for robot actions, a servo motor for camera adjustment, and a buzzer to diversify robot actions. The communication between CAROBO and LA-RCS was configured to utilize ROS (Robot Operating System) communication protocols.
\begin{figure}[htbp]
    \centering
    \includegraphics[width=\textwidth]{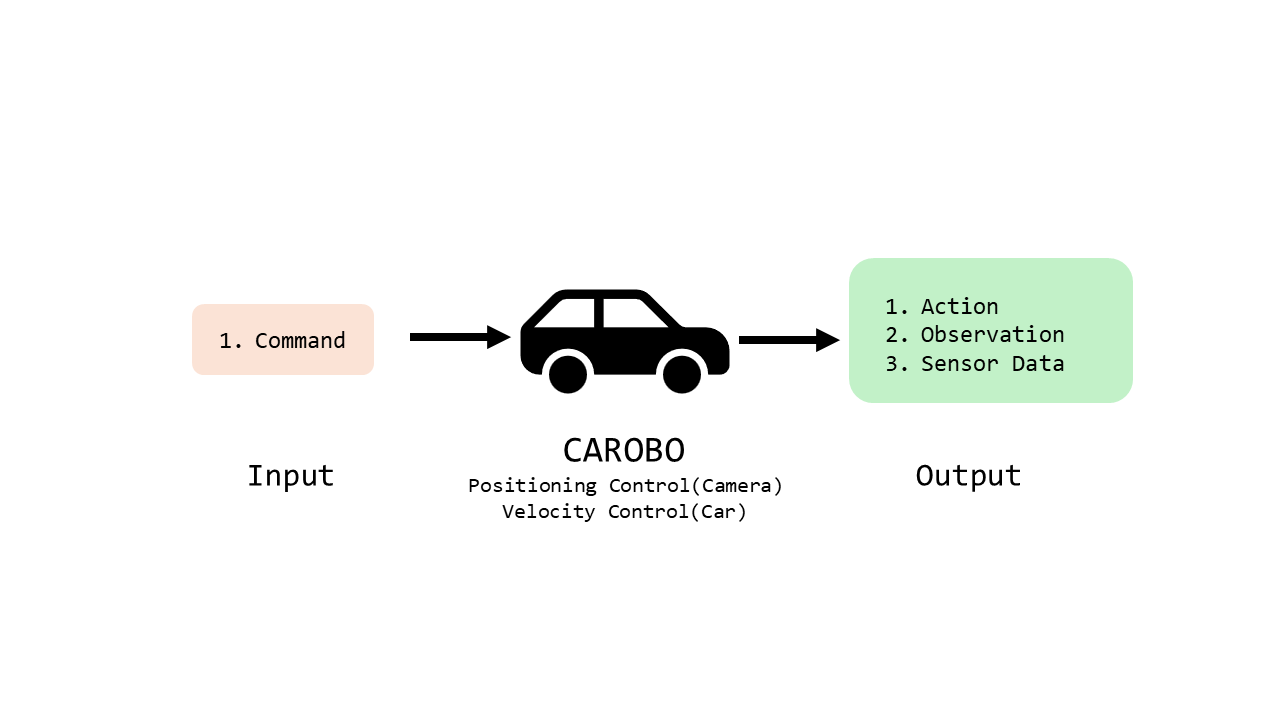}
    \caption{An illustration of the CAROBO.}
    \label{CAROBO}
\end{figure}

\subsubsection{CAROBO Input}
\begin{itemize}
    \item \textbf{Control Function (Command)}: Action to be performed by the robot.
\end{itemize}

The Control Functions received by the robot are predefined in CAROBO. The following Functions exist in CAROBO:
\begin{itemize}
    \item \textbf{car forward}: Command for the robot to move forward. It has a feature of turning right when it gets too close to an object in front.
    \item \textbf{car back}: Command for the robot to move backward.
    \item \textbf{car left}: Command for the robot to turn left.
    \item \textbf{car right}: Command for the robot to turn right.
    \item \textbf{camera move}: Command to adjust the angle of the robot's camera.
\end{itemize}

The App Agent assesses the situation and selects one of these Control Functions (Commands) to execute robot control. Upon receiving this input, CAROBO produces the following outputs:

\subsubsection{CAROBO Output}
\begin{itemize}
    \item \textbf{Action}: The robot performs the action according to the Command.
    \item \textbf{Vision Data}: Visual data for confirming the robot's next action or action termination.
    \item \textbf{Sensor Data}: Data for confirming action or action termination based on the distance between the robot and objects.
\end{itemize}

CAROBO executes the Control Function received from the App Agent. Vision Data and Sensor Data are transmitted to the App Agent, becoming data for the App Agent to fulfill the User Request.

\section{Evaluation}

To evaluate the performance of LA-RCS, we assessed the success and failure of User Requests across four domains. These domains were object detection, command execution, obstacle navigation, and situation awareness. For each domain, five requests were performed. The models used as Agents were GPT-4-Turbo and GPT-4o. User Requests were inputted, and the robot and LA-RCS were configured to share data through ROS communication. The number of steps represents the count of commands issued by the App Agent to the robot to fulfill the User Request, indicating how many iterations the App Agent performed. Failure criteria included meaningless movements beyond a certain number of steps or LA-RCS determining completion when the User Request was not actually fulfilled.

\subsection{Benchmark}
To comprehensively assess the performance of LA-RCS, we developed a benchmark consisting of 20 User Requests. These requests were designed across four domains: object detection, command execution, obstacle navigation, and situational awareness. This approach ensures diversity and comprehensiveness in the evaluation. Five User Requests were designed for each domain, resulting in a total of 20 requests.

For the Agent component, we selected GPT-4-Turbo and GPT-4o as the models. We evaluate each User Request from two perspectives: success and failure. Success indicates that LA-RCS successfully completed the User Request. Failure is determined when LA-RCS incorrectly judges a request as completed when it has not been, or when it exhibits repetitive, meaningless movements for a certain number of steps. A step represents an action taken by LA-RCS to complete the request.
\subsection{Performance Evaluation}
Tables \ref{Object Detection Results}, \ref{Command Execution Results}, \ref{Obstacle Navigation Results}, and \ref{Situation Awareness Results} present detailed success and failure rates for User Requests in the four domains: object detection, command execution, obstacle navigation, and situational awareness, respectively.
\subsubsection{Object Detection}

\begin{table}[htbp]
\centering
\begin{tabular}{|c|>{\centering\arraybackslash}p{6cm}|>{\centering\arraybackslash}p{3cm}|>{\centering\arraybackslash}p{3cm}|}
\hline
 & Request - Object Detection & GPT-4-Turbo & GPT-4o \\ \hline
1 & Move around and search for the location of the refrigerator. & Success - Step : 6 & Success - Step : 9 \\ \hline
2 & Look for any people in the immediate vicinity. & Success - Step : 8 & Success - Step : 4 \\ \hline
3 & Move around and search for any people in the surrounding area. & Failure - Step : 20 & Success - Step : 11 \\ \hline
4 & If a rectangular object is seen, move towards it and activate the buzzer. & Success - Step : 4 & Success - Step : 2 \\ \hline
5 & Find yellow obstacles and tell me what is written on them. & Success - Step : 0 & Success - Step : 0 \\ \hline
\end{tabular}
\caption{Object Detection Results}
\label{Object Detection Results}
\end{table}

User Requests for the object detection domain are presented in Table \ref{Object Detection Results}. In this domain, the Agent composed of GPT-4-Turbo successfully completed 4 out of 5 tasks, demonstrating an 80\% success rate, while the Agent composed of GPT-4o achieved a 100\% success rate by completing all 5 tasks.

\subsubsection{Command Execution}

\begin{table}[htbp]
\centering
\begin{tabular}{|c|>{\centering\arraybackslash}p{6cm}|>{\centering\arraybackslash}p{3cm}|>{\centering\arraybackslash}p{3cm}|}
\hline
 & Request - Command Execution & GPT-4-Turbo & GPT-4o \\ \hline
1 & Rotate in a 0.6-meter square shape. & Failure - Step : 5 & Success - Step : 9 \\ \hline
2 & Lift your head, identify a human face, and describe it. & Success - Step : 3 & Success - Step : 3 \\ \hline
3 & Rotate twice on the spot & Success - Step : 1 & Success - Step : 8 \\ \hline
4 & Move in a zigzag pattern at a 30-degree angle for a distance of 2 meters. & Success - Step : 5 & Success - Step : 10 \\ \hline
5 & Move backwards 0.4 meters and sound the buzzer, repeat this 5 times. & Success - Step : 12 & Success - Step : 12 \\ \hline
\end{tabular}
\caption{Command Execution Results}
\label{Command Execution Results}
\end{table}

User Requests for the command execution domain are presented in Table \ref{Command Execution Results}. In this domain, the Agent composed of GPT-4-Turbo successfully completed 4 out of 5 tasks, demonstrating an 80\% success rate, while the Agent composed of GPT-4o achieved a 100\% success rate by completing all 5 tasks.

\subsubsection{Obstacle Navigation}
\begin{table}[htbp]
\centering
\begin{tabular}{|c|>{\centering\arraybackslash}p{6cm}|>{\centering\arraybackslash}p{3cm}|>{\centering\arraybackslash}p{3cm}|}
\hline
 & Request - Obstacle Navigation & GPT-4-Turbo & GPT-4o \\ \hline
1 & Move forward avoiding obstacles & Failure - Step : 16 & Success - Step : 7 \\ \hline
2 & Move forward for a total of 2 meters, turning right and activating the buzzer when obstacles appear. & Success - Step : 8 & Success - Step : 5 \\ \hline
3 & Move to find a bosch box while avoiding obstacles & Failure - Step : 15 & Success - Step : 8 \\ \hline
4 & Rotate once while observing the surroundings, then move 1 meter in a direction without obstacles & Failure - Step : 5 & Failure - Step : 15 \\ \hline
5 & After observing obstacles in front, move behind the observed object, stop, and activate the buzzer. & Failure - Step : 3 & Failure - Step : 5 \\ \hline
\end{tabular}
\caption{Obstacle Navigation Results}
\label{Obstacle Navigation Results}

\end{table}

User Requests for the obstacle navigation domain are presented in Table \ref{Obstacle Navigation Results}. In this domain, the Agent composed of GPT-4-Turbo successfully completed 1 out of 5 tasks, demonstrating a 20\% success rate, while the Agent composed of GPT-4o successfully completed 4 out of 5 tasks, demonstrating an 60\% success rate.

\subsubsection{Situation Awareness}

\begin{table}[htbp]
\centering
\begin{tabular}{|c|>{\centering\arraybackslash}p{6cm}|>{\centering\arraybackslash}p{3cm}|>{\centering\arraybackslash}p{3cm}|}
\hline
 & Request - Situation Awareness & GPT-4-Turbo & GPT-4o \\ \hline
1 & Describe the features of the object in front & Success - Step : 0 & Success - Step : 0 \\ \hline
2 & Detect the surroundings and describe only navy-colored objects in Korean & Success - Step : 6 & Success - Step : 7 \\ \hline
3 & Move forward 0.5 meters, observe the surroundings, and tell me the name of the box. & Success - Step : 3 & Success - Step : 7 \\ \hline
4 & After moving 2 meters, if there is a paper in front, print out what is written on it. & Failure - Step : 3 & Success - Step : 10 \\ \hline
5 & From the paper observed in front, tell me the contact information for the Ministry of Science, ICT and Future Planning. & Failure - Step : 4 & Success - Step : 0 \\ \hline
\end{tabular}
\caption{Situation Awareness Results}
\label{Situation Awareness Results}
\end{table}

User Requests for the situation awareness domain are presented in Table \ref{Situation Awareness Results}. In this domain, the Agent composed of GPT-4-Turbo successfully completed 3 out of 5 tasks, demonstrating an 60\% success rate, while the Agent composed of GPT-4o achieved a 100\% success rate by completing all 5 tasks.

One notable observation is that the obstacle navigation area in Table \ref{Obstacle Navigation Results} shows relatively lower success rates compared to other areas. The obstacle navigation domain typically requires more extensive use of the robot's sensor data compared to other domains. We attribute the difficulties in executing Requests in the obstacle navigation area to two factors: the limited directionality of the sensors attached to the robot and the constraint of executing only one Control Function (Command) per Step. Nevertheless, the Agent composed of GPT-4o demonstrates an overall success rate of 95\%, solidifying its position as an effective controller for robotic systems.

\section{Case Study}
Herein, we present two illustrative cases demonstrating how the LA-RCS effectively processes user requests. These examples highlight the system's capabilities and provide insight into its operational mechanisms.

\subsection{Exploration and Object Detection in the Surrounding Environment}

\begin{figure}[htbp]
    \centering
    \includegraphics[width=\textwidth]{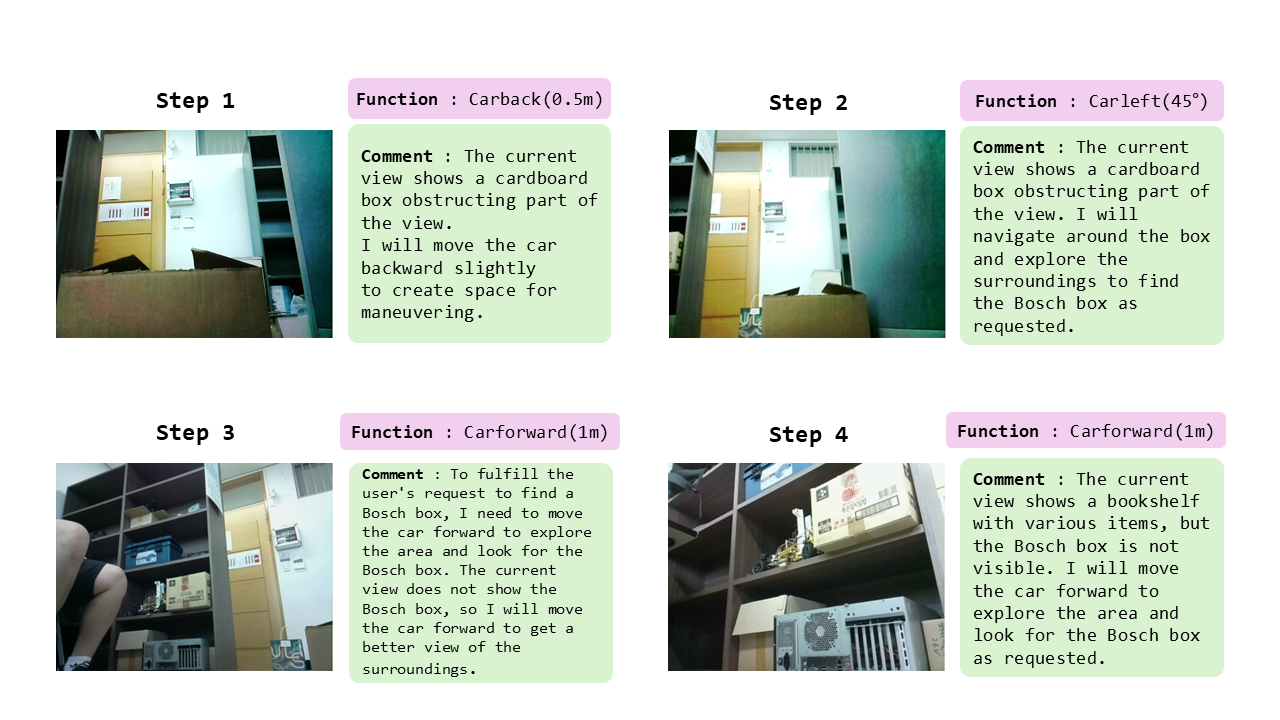}
    \label{Casestudy1_1}
    \includegraphics[width=\textwidth]{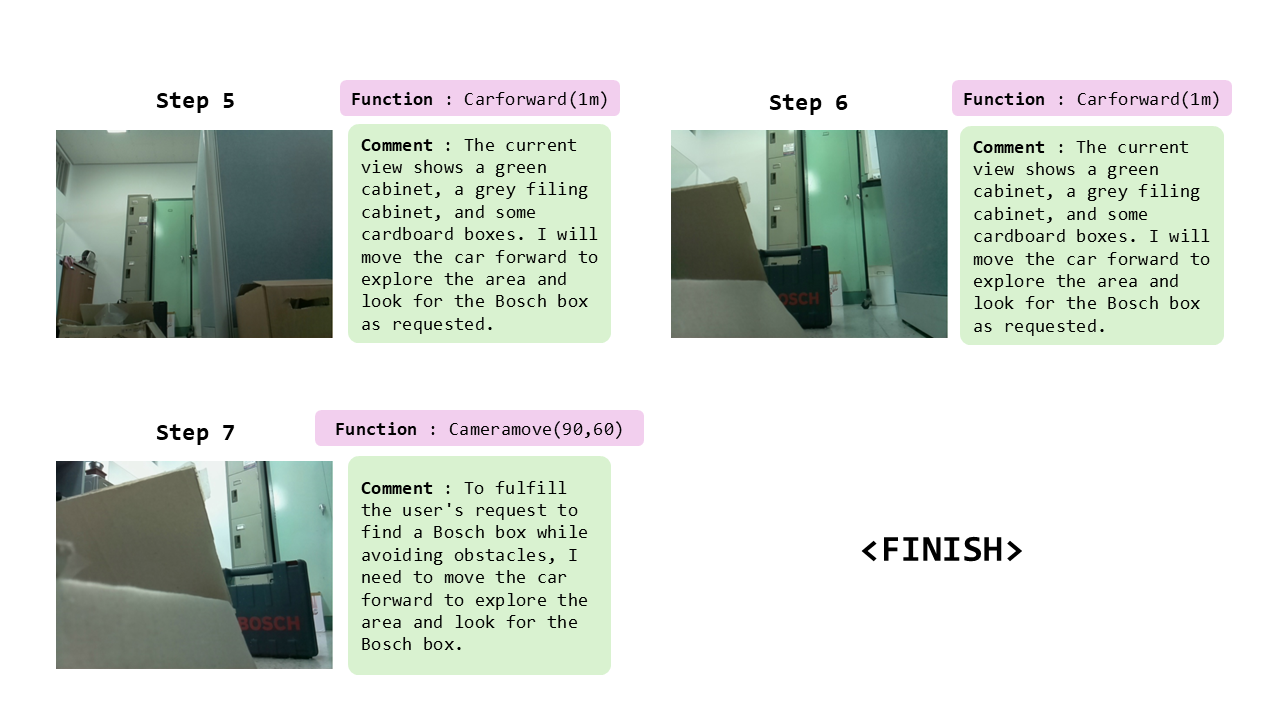}
    \label{Casestudy1_2}
    
    \caption{An detailed example of completing user request: "Find the box that says Bosch.”}
    \label{Casestudy1}
\end{figure}

In Figure \ref{Casestudy1}, we tasked LA-RCS with the request:"Find the box that says Bosch.”. This scenario involves placing various objects around the environment. For the LA-RCS to execute a request, it must continuously assess the current space, constantly search for the given objects, and upon discovering the specified target, complete the challenging. request.

To navigate such a situation, the LA-RCS autonomously determines the need to execute the Carback function to secure its view. Based on the acquired visual information, it analyzes where to move next and proceeds forward. Subsequently, upon identifying an object that is expected to be the "Bosch Box," it raises its head to confirm the match and completes the request.

\subsection{Obstacle Passage}
\begin{figure}[htbp]
    \centering  
    \includegraphics[width=\textwidth]{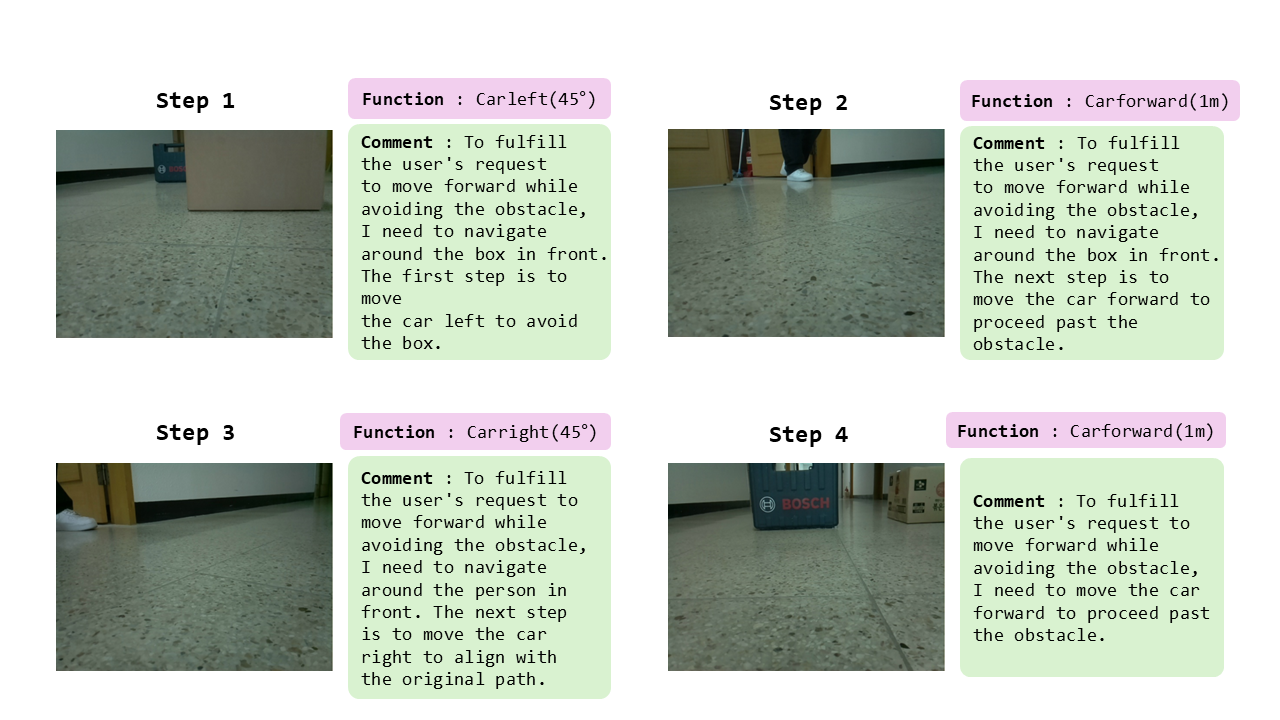}
    \label{CaseStudy2_1}
    \includegraphics[width=\textwidth]{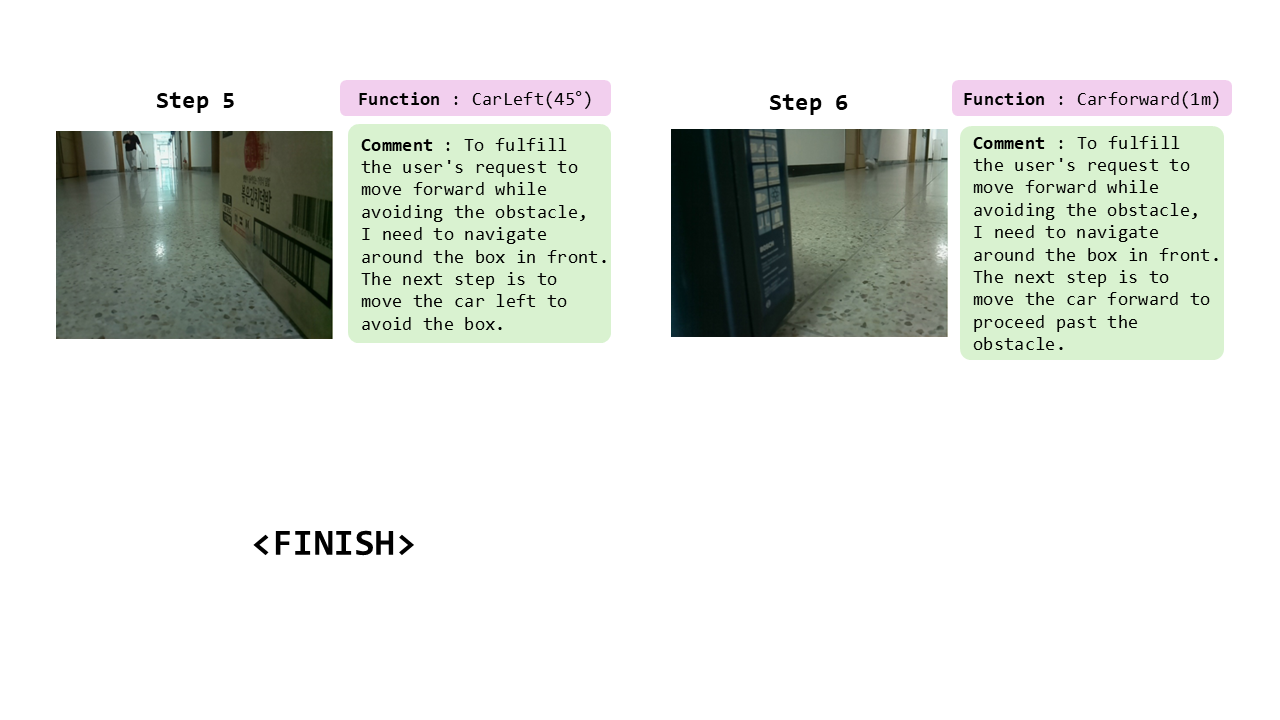}
    \label{CaseStudy2_2}
    
    \caption{An detailed example of completing user request: "Move forward avoiding obstacles.”}
    \label{CaseStudy2}
\end{figure}

In Figure \ref{CaseStudy2}, we tasked LA-RCS with the request:"Move forward avoiding obstacles.”.

This scenario involves navigating various obstacles. To execute a request, the LA-RCS must continuously assess the current space and determine how to move autonomously, making it a challenging task. To handle such a situation, the LA-RCS observes the external environment and calculates its actions to move forward.

These two outcomes demonstrate that the LA-RCS can analyze how to fulfill the user's request, plan accordingly, move the robot, and continuously control its movements while observing the external environment.
\section{Conclusion \& Future Work}
Our research explored the LLM Agent system, where robots autonomously perform given commands through observation and feedback. The LA-RCS system is based on a Dual-Agent Framework, separating the macro-level Task Planning Agent and the Agent that physically executes the tasks, engaging in conversation to accomplish the given commands.

This approach enables the agent to independently resolve unfamiliar situations, decompose complex tasks into manageable parts, and provide feedback to handle unexpected anomalies. The refinement of this method suggests the potential for developing a system capable of executing commands with minimal human intervention in various scenarios.

However, our study utilized API-based Agent systems like GPT-4, Gemini, and Claude, resulting in considerable time consumption for command execution. The necessity for repeated prompt inferences further contributed to the time delays. Additionally, the limited command execution capabilities of the robot restricted the full accomplishment of some tasks.

In future work, we aim to develop a proprietary backbone LLM to maximize the AutoRegressive performance, ensuring faster processing speeds. We plan to create task-specific LLM models to advance our research. Furthermore, we will evaluate the efficacy of our system in more sophisticated environments beyond the current setup, which primarily involves vision camera position control and vehicle speed control.

\end{document}